\documentclass[12pt,onecolumn,letterpaper]{article}

\usepackage{iccv}
\usepackage{times}
\usepackage{epsfig}
\usepackage{graphicx}
\usepackage{amsmath}
\usepackage{amssymb}

\usepackage{multirow}
\usepackage{soul}
\usepackage{subcaption}
\usepackage{enumitem}

\usepackage{appendix}
\usepackage{etoolbox}
\usepackage{lipsum}
\usepackage[dvipsnames]{xcolor}

\definecolor{mygray}{gray}{0.6}

\newcommand\blfootnote[1]{%
  \begingroup
  \renewcommand\thefootnote{}\thanks{#1}%
  \endgroup
}

\usepackage[pagebackref=true,breaklinks=true,letterpaper=true,colorlinks,bookmarks=false]{hyperref}

\iccvfinalcopy 

\def\iccvPaperID{1259} 

\ificcvfinal\fi

\begin{document}

\title{Text Prior Guided Scene Text Image Super-resolution}

\author{Jianqi Ma, Shi Guo, Lei Zhang\\
Dept. of Computing, The Hong Kong Polytechnic University\\
{\tt\small \{csjma, cssguo, cslzhang\}@comp.polyu.edu.hk}
\blfootnote{This work is supported by the Hong Kong RGC RIF grant (R5001-18).}


}

\maketitle
\ificcvfinal\thispagestyle{empty}\fi

\begin{abstract}
      Scene text image super-resolution (STISR) aims to improve the resolution and visual quality of low-resolution (LR) scene text images, and consequently boost the performance of text recognition. However, most of existing STISR methods regard text images as natural scene images, ignoring the categorical information of text. In this paper, we make an inspiring attempt to embed categorical text prior into STISR model training. Specifically, we adopt the character probability sequence as the text prior, which can be obtained conveniently from a text recognition model. The text prior provides categorical  guidance to recover high-resolution (HR) text images. On the other hand, the reconstructed HR image can refine the text prior in return. Finally, we present a multi-stage text prior guided super-resolution (TPGSR) framework for STISR. Our experiments on the benchmark TextZoom dataset show that TPGSR can not only effectively improve the visual quality of scene text images, but also significantly improve the text recognition accuracy over existing STISR methods. Our model trained on TextZoom also demonstrates certain generalization capability to the LR images in other datasets.
\end{abstract}

\section{Introduction}

\label{intro}

Scene text image recognition aims to recognize the text characters from the input image, which is an important computer vision task that involves text information processing. It has been widely used in text retrieval~\cite{karaoglu2016words}, sign recognition~\cite{fang2004automatic}, license plate recognition~\cite{montazzolli2018license} and other scene-text-based image understanding tasks~\cite{biten2019scene,mathew2020docvqa}. However, due to the various issues such as low sensor resolution, blurring, poor illumination, etc., the quality of captured scene text images may not be good enough, which brings many difficulties to scene text recognition in practice. In particular, scene text recognition from low-resolution (LR) images remains a challenging problem. 

\begin{figure}[t]
  \centering
  \includegraphics[width=0.8\linewidth]{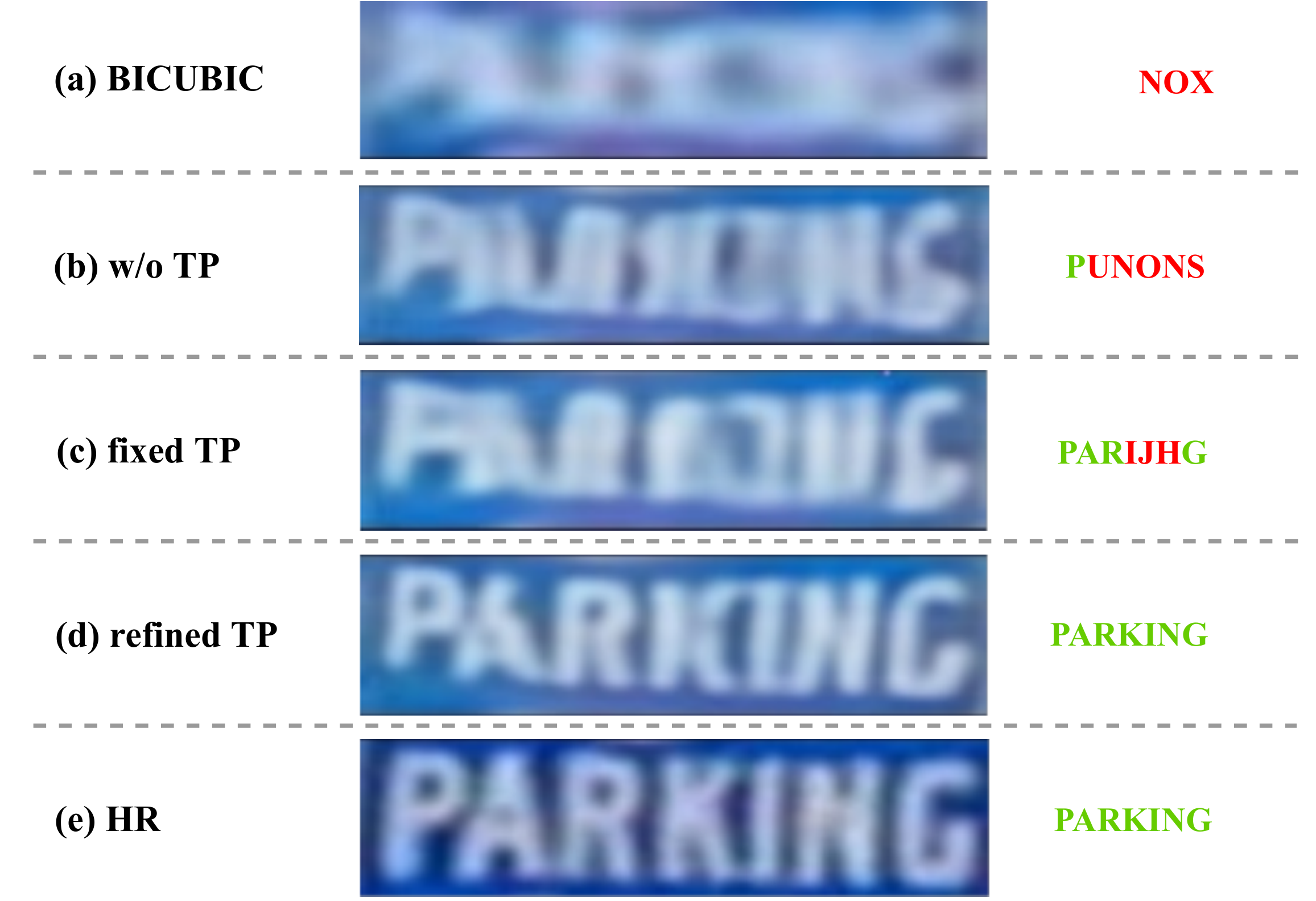}
  \caption{Comparison of super-resolution results generated from TSRN~\cite{wang2020scene} with and without (w/o) our text prior (TP). The right column shows the text recognition results.}
  \label{fig:comparison_TP}
\vspace{-0.3cm}
\end{figure}

In recent years, single image super-resolution (SISR) techniques~~\cite{dong2016accelerating,kim2016accurate,lai2017deep,ledig2017photo,lim2017enhanced,zhang2018residual,zhang2018image} have achieved significant progress owing to the rapid development of deep neural networks~\cite{krizhevsky2017imagenet,simonyan2014very,he2016deep}. Inspired by the success of SISR, researchers have started to investigate scene text image super-resolution (STISR) to improve the quality of LR text images and hence improve the text recognition accuracy. Tran \etal~\cite{tran2019deep} adapted LapSRN~\cite{lai2017deep} to STISR and significantly improved the text content details. To obtain more realistic STISR results, B{\'\i}lkov{\'a} \etal~\cite{bilkova2020perceptual} and Wang \etal~\cite{wang2019textsr} employed GAN based networks with CTC loss~\cite{graves2006connectionist} and text perceptual losses. In these methods, the LR images are synthesized (\eg, bicubic down-sampling) from high resolution (HR) images for SR model learning, while the image degradation process in real-world LR images can be much more complex. Recently, Wang \etal~\cite{wang2020scene} collected a real-world STISR dataset, namely TextZoom, where LR-HR image pairs captured by zooming lens are provided. Wang \etal also proposed a TSRN model for STISR, achieving state-of-the-art performance ~\cite{wang2020scene}.

The existing STISR methods~\cite{tran2019deep,wang2019textsr,bilkova2020perceptual}, however, mostly treat scene text images as natural scene images to perform super-resolution, ignoring the important categorical information brought by the text content in the image. As shown in Fig.~\ref{fig:comparison_TP}(b), the result by TSRN~\cite{wang2020scene} is much better than the simple bicubic model (Fig.~\ref{fig:comparison_TP}(a)), but it is still hard to tell the characters therein. Based on the observation that semantic information can help to recover the shape and texture of objects~\cite{wang2018recovering}, in this paper we propose a new STISR method, namely text prior guided super-resolution (TPGSR), by introducing the categorical text prior information into the model learning process. Unlike the face segmentation prior used in \cite{chen2018fsrnet} and the semantic segmentation used in \cite{wang2018recovering}, the text character segmentation is hard to obtain since there are few datasets containing annotations of fine character segmentation masks. We instead employ a text recognition model (\eg, CRNN~\cite{shi2016end}) to extract the probability sequence of the text as the categorical prior of the given LR scene text image, and embed it into the super-resolution network learning process to guide the reconstruction of HR image. As can be seen in Fig.~\ref{fig:comparison_TP}(c), the text prior information can indeed improve much the STISR results, making the text characters much more readable. On the other hand, the reconstructed HR text image can be used to refine the text prior, and consequently a multi-stage TPGSR framework can be built for effective STISR. Fig.~\ref{fig:comparison_TP}(d) shows the super-resolved text image by using the refined text prior, where the text can be clearly and correctly recognized. 
The major contributions of our work are as follows:
\begin{itemize}[topsep=2pt]
\setlength{\topsep}{0pt}
\setlength{\itemsep}{0pt}
\setlength{\parsep}{0pt}
\setlength{\parskip}{0pt}

\item We introduce the text recognition categorical probability sequence as the prior into the STISR task, and validate its effectiveness to improve the visual quality of text content and recognition accuracy of scene text images. 

\item We propose the refinements of the text recognition categorical prior without using extra supervision of text labels except for the real HR image: refining recurrently by the estimated HR image and by fine-tuning the text prior generator with our proposed TP Loss. With such refinement, the text prior and the super-resolved text image can be jointly enhanced under our TPGSR framework. 

\item By improving the image quality of LR text images, the proposed TPGSR improves the text recognition performance on TextZoom for different text recognition models by a large margin and demonstrates good generalization performance to other recognition datasets.
\end{itemize}

\section{Related Works}

\textbf{Single Image Super Resolution (SISR).} Aiming at estimating a high-resolution (HR) image from its low-resolution (LR) counterpart, SISR is a highly ill-posed problem. In the past, handcrafted image priors are commonly used to regularize the SISR model to improve the image quality. In recent years, the training of deep neural networks (DNNs) has dominated the research of SISR. The pioneer work SRCNN~\cite{dong2015image} learns a three-layer convolutional neural network (CNN) for the SISR task. Later on, many deeper CNN models have been proposed to improve the SISR performance, \eg, deep residual block~\cite{lim2017enhanced}, Laplacian pyramid structure~\cite{lai2017deep}, densely connected network~\cite{zhang2018residual} and channel attention mechanism~\cite{zhang2018image}. The PSNR and SSIM~\cite{wang2004image} losses are widely used in those works to train the SISR model. In order to produce perceptually-realistic SISR results, SRGAN~\cite{ledig2017photo} employs a generative adversarial network (GAN) to synthesize image details. SFT-GAN~\cite{wang2018recovering} and FSRNet~\cite{chen2018fsrnet} utilizes the GAN loss and semantic segmentation to generate visually pleasing HR image. 

\textbf{Scene Text Image Super Resolution (STISR)}. Different from the general purpose SISR that works on natural scene images, STISR focuses on text images, aiming to improve the readability of texts by improving their visual quality. Intuitively, those methods for SISR can be directly adopted for STISR. In~\cite{dong2015boosting}, Dong \etal extended SRCNN~\cite{dong2015image} to text images, and obtained the best performance in ICDAR 2015 competition ~\cite{peyrard2015icdar2015}. PlugNet~\cite{mouplugnet} employs a light-weight pluggable super-resolution unit to deal with LR images in feature domain. TextSR~\cite{wang2019textsr} utilizes the text recognition loss and text perceptual loss to generate the desired HR images for text recognition. To improve the performance of STISR on real-world scene text images, Wang \etal \cite{wang2020scene} built a real-world STISR image dataset, namely TextZoom, where the LR and HR text image pairs were cropped from real-world SISR datasets~\cite{zhang2019zoom,cai2019toward}. They also proposed a TSRN~\cite{wang2020scene} method to use the central alignment module and sequential residual block to exploit the semantic information in internal features. SCGAN~\cite{xu2017learning} employs a multi-class GAN loss as supervision to equip the model with ability to generate more distinguishable face and text images. Further by progressively adopting the high-frequency information derived from the text image, Quan \etal~\cite{quan2020collaborative} proposed a multi-stage model for recovering the blurry text images.
Different from the above methods, we propose employ the text recognition prior to guide the STISR model recovering better quality text images.

\begin{figure}[t]
\setlength{\abovecaptionskip}{0.3cm}
  \centering
  \includegraphics[width=0.8\linewidth]{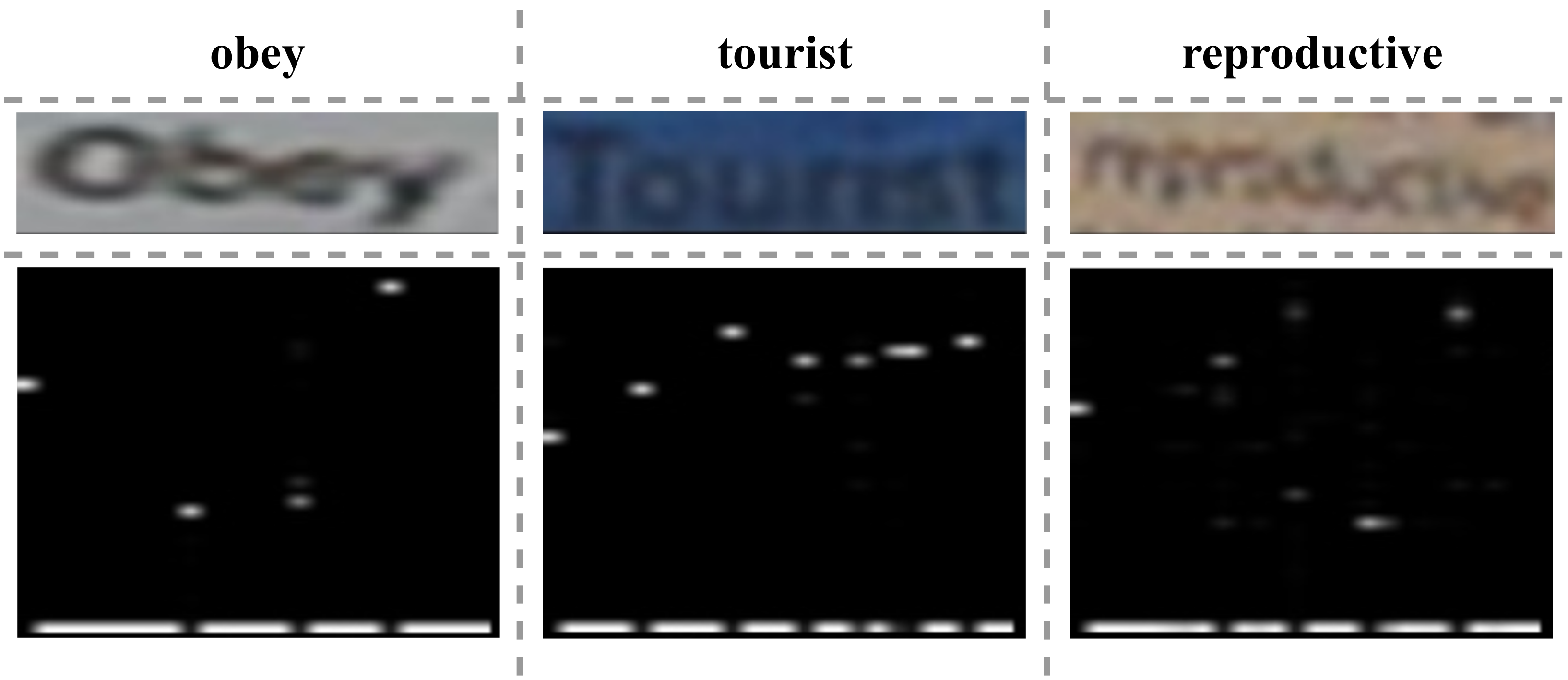}
  \caption{Visualization of the text prior (TP) of some text images. The top, middle and bottom rows present the text labels, input text images and their TPs, respectively.}
  \label{fig:TP_examples}
\vspace{-0.3cm}
\end{figure}

\textbf{Scene Text Recognition.} In the early stage of deep learning based scene text recognition, researchers intended to solve the problem in a bottom-up manner \cite{jaderberg2014deep,he2015reading}, \ie, extracting the text from characters into words. Some other approaches recognize the text in a top-down fashion \cite{jaderberg2016reading}, \ie, regarding the text image as a whole and performing a word-level classification. Taking text recognition as an image-to-sequence problem, CRNN~\cite{shi2016end} employs the CNN to extract image feature and uses the recurrent neural networks to model the semantic information of image features. Trained with the CTC~\cite{graves2006connectionist} loss, the predicted sequence can be more accurately aligned with the target sequence \cite{liu2016star}. Recently, attention-based methods thrive due to the improvement in text recognition benchmarks and the robustness to various shapes of text images \cite{cheng2017focusing,cheng2018aon}. In our method, we adopt CRNN as the text prior generator to generate categorical text priors for STISR model training. It is shown that such text priors can significantly improve the perceptual quality of super-resolved text images and consequently boost the text recognition performance. 

\section{Methodology}
In this section, we will first explain what the text prior (TP) is, and then introduce the text prior guided super-resolution (TPGSR) network in detail, followed by the design of loss function.

\subsection{Text Prior}
\label{sectp}
In this paper, the TP is defined as the deep categorical representation of a scene text image generated by some text recognition models. The TP is then used as guidance information to encourage our TPGSR network to produce high quality scene text images, which are favorable to both visual perception and scene text recognition. 

Specifically, we choose the classic text recognition model CRNN~\cite{shi2016end} to be the TP Generator. CRNN uses several convolution layers to extract the text features and five max pooling layers to down-sample the features into a feature sequence. The TP is then defined as the categorical probability prediction by CRNN, which is a sequence of $|A|$-dimensional probability vectors where $|A|$ denotes the number of characters learned by CRNN.
%
%
Fig.~\ref{fig:TP_examples} visualizes the TP of some scene text images, where the horizontal axis represents the sequence in left-to-right order and the vertical axis represents the categories in reverse alphabet order (\eg, 'Z' to 'A'). In the visualization, the lighter the spot is, the higher the probability of this category will be. By using the TP as guidance, our TPGSR model can recover visually more pleasing HR images with higher text recognition accuracy, as we illustrated in Fig.~\ref{fig:comparison_TP}.

\begin{figure*}[t]
  \centering
  \includegraphics[width=\linewidth]{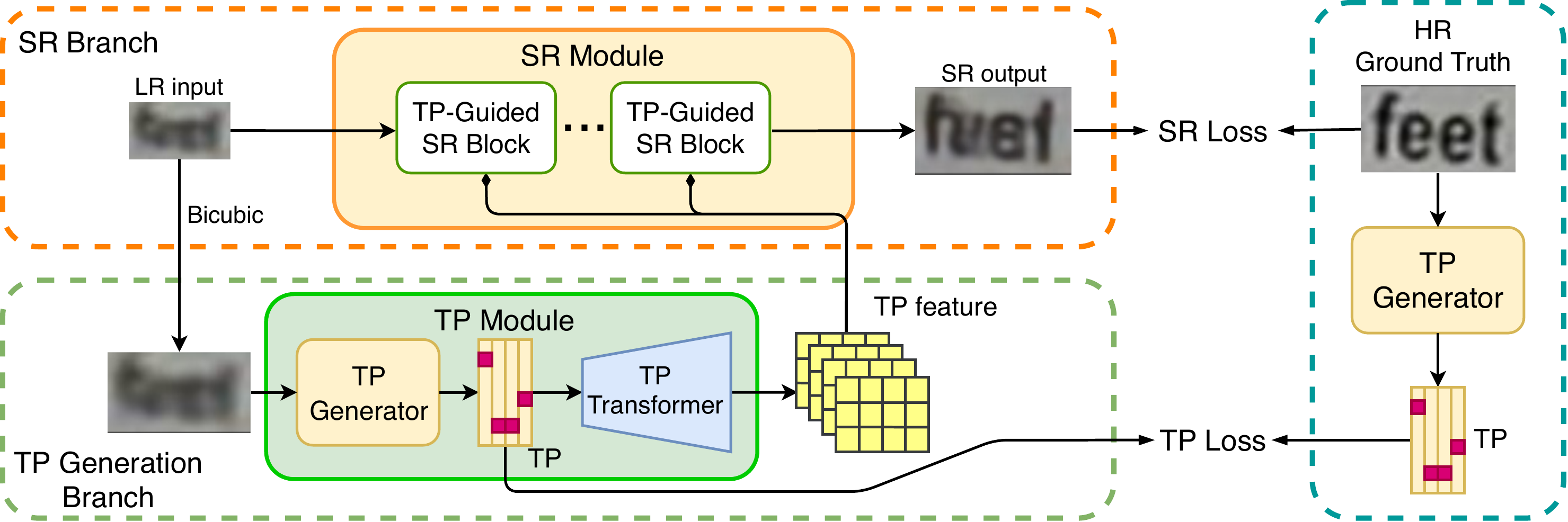}
  \caption{Our proposed TPGSR framework, which consists of a Text Prior Generation Branch and a Super-resolution (SR) Branch. Accordingly, TP loss and SR loss are employed to train the whole network.}
 \label{fig:TPG}
\vspace{-0.3cm}
\end{figure*}

\subsection{The Architecture of TPGSR}
\label{SecArcTPGSR}
By introducing TP into the STISR process, the main architecture of our TPGSR is illustrated in Fig.~\ref{fig:TPG}. Our TPGSR network has two branches: a TP generation branch and a super-resolution (SR) branch. First, the TP branch intakes the LR image to generate the TP feature. Then, the SR branch intakes the LR image and the TP feature to estimate the HR image. In the following, we introduce the two branches in detail. 

\noindent\textbf{TP generation branch.} This branch uses the input LR image to generate the TP feature and passes the TP feature to the SR branch as guidance for more accurate STISR results. The key component of this branch is the TP Module, which consists of a TP Generator and a TP Transformer. As mentioned in Section~\ref{sectp}, the TP generated by TP Generator is a probability sequence, whose size may not match the image feature map in the SR branch. To solve this problem, we employ a TP Transformer to transform the TP sequence into a feature map.

Specifically, the input LR image is first resized to match the input of TP Generator by bicubic interpolator, and then passed to the TP Generator to generate a TP matrix whose width is $L$. Each position of the TP is a vector of size $|A|$, which is the number of categories of alphabet $A$ adopted in CRNN. To align the size of TP feature with the size of image feature, we pass the TP feature to the TP Transformer. 
The TP Transformer consists of 4 Deconv blocks, each of which consists of a deconvolution layer, a BN layer and a ReLU layer. For an input TP matrix with width $L$ and height $|A|$, the output of TP Transformer will be a feature map with recovered spatial dimension and
channel $32$ after three deconvolution layers with stride $(2,2)$ and one deconvolution layer with stride $(2,1)$. The kernel size of the all deconvolution layers is $3\times3$.

\noindent\textbf{SR branch} The SR branch aims to reproduce an HR text image from the input LR image and TP guidance feature. It mainly contains an SR Module. Please note that most of the SR blocks in existing SISR and STISR methods~\cite{kim2016accurate,lai2017deep,ledig2017photo,lim2017enhanced,zhang2018residual,tran2019deep,wang2019textsr,wang2020scene} can be adopted as our SR Module in couple with our TP guidance features. Considering that these SR blocks, such as the residual block in SRResNet~\cite{ledig2017photo} and the SRB in TSRN~\cite{wang2020scene}, only take the image features as input, we need to modify them in order to embed the TP features. We call our modified SR blocks as TP-Guided SR Blocks. 

\begin{figure}[t]
  \centering
  \includegraphics[width=0.8\linewidth]{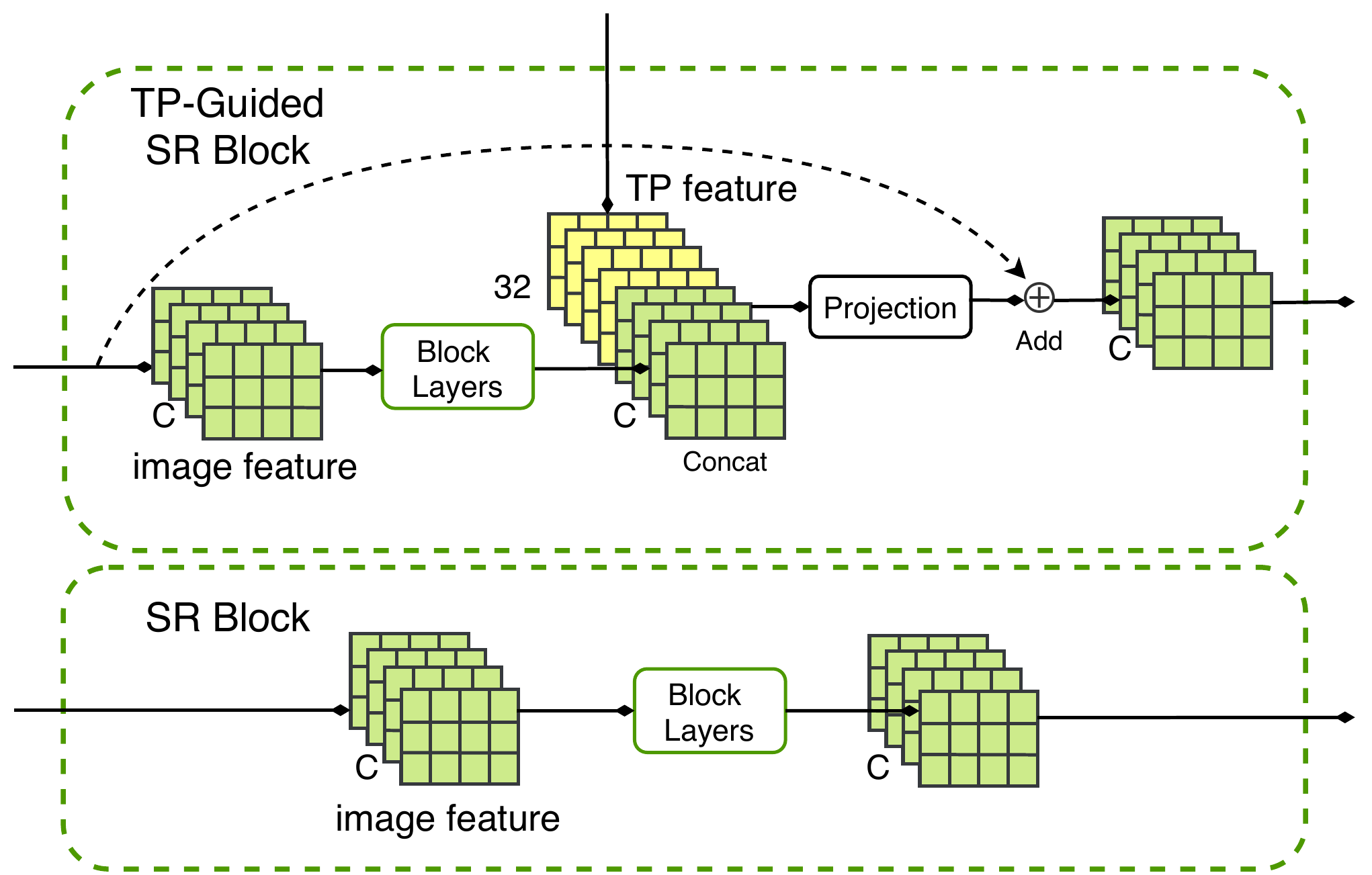}
  \caption{Comparison of our TP-Guided SR Block and a common SR Block. In each block, the channel numbers of image features and TP features are $C$ and $32$, repectively.}
  \label{fig:TPB}
\end{figure}

The difference between previous SR Blocks and our TP-Guided SR Block is illustrated in Fig.~\ref{fig:TPB}. To embed the TP features into the SR Block, we concatenate them to the image features along the channel dimension. Before the concatenation, we align the spatial size of TP features to that of the image features by bicubic interpolation. Suppose that the channel number of image features is $C$, then the concatenated features of $C+32$ channels will go through a projection layer to reduce the channel number back to $C$. We simply use a $1\times1$ kernel convolution to perform this projection. The output of projection layer is fused with the input image feature by addition. With several such TP-Guided SR Blocks, the SR branch will output the estimated HR image, as in those previous super-resolution models.


\subsection{Multi-stage Refinement}
\begin{figure}[t]
  \centering
  \includegraphics[width=0.8\linewidth]{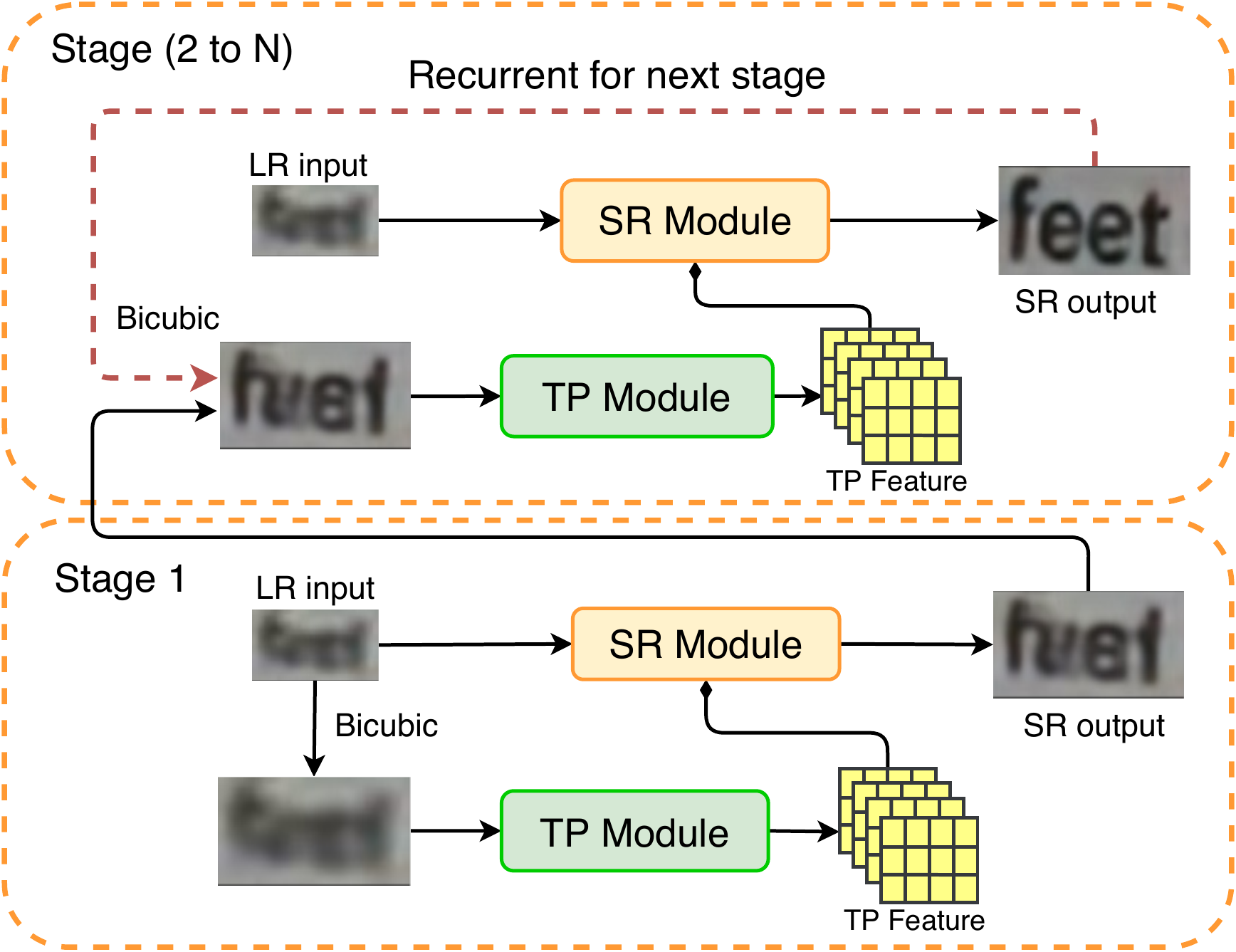}
  \caption{Illustration of multi-stage TPGSR. The super-resolution output of one stage will be the text image input of next stage.}
  \label{fig:MultiStage}
\vspace{-0.3cm}
\end{figure}

With the TPGSR framework described in Section~\ref{SecArcTPGSR}, we can super-resolve an LR image to a better quality HR image with the help of TP features extracted from the LR input. One intuitive question is, if we extract the TP features from the super-resolved HR image, can we use those better quality TP features to further improve the super-resolution results? Actually, multi-stage refinement has been widely adopted in many computer vision tasks such as object detection~\cite{cai2018cascade} and instance segmentation~\cite{chen2019hybrid} to improve the prediction quality progressively. Therefore, we extend our one-stage TPGSR model to a multi-stage learning framework by passing the estimated HR text image in one stage to the TP Generator in next stage. The multi-stage TPGSR framework is illustrated in Fig.~\ref{fig:MultiStage}. In the $1^{\text{st}}$ stage, the TP Module accepts the bicubically interpolated LR image as input, while in the following stages, the TP Module accepts the HR image output from the SR Module in previous stage as input for refinement. As we will show in the ablation study in Section~\ref{secablation}, both the quality of estimated HR text image and the text recognition accuracy can be progressively improved by this multi-stage refinement.

\subsection{Training Loss} 
There are two types of loss functions in our TPGSR framework, one for the SR branch and another one for the TP generation branch. For the SR branch, the loss is similar to that in many previous SISR methods~\cite{lai2017deep,lim2017enhanced,zhang2018residual}. Denote by $\hat{I}_H$ the estimate HR image from the LR input and by $I_H$ the ground-truth HR image, the loss for the SR branch, denoted by $L_S$, can be commonly defined as the $L_1$ norm distance between $\hat{I}_H$ and $I_H$, \ie, $L_S=|\hat{I}_H - I_H|$.

Different from the many SISR works~\cite{lai2017deep,lim2017enhanced,zhang2018residual} as well as the STISR works~\cite{wang2019textsr,wang2020scene}, in TPGSR we have loss functions specifically designed for the TP generation branch, which is crucial to improve the text image quality and text recognition. The TP sequence generated by the TP Generator has significant impact on the final SR results. If the TP is correct (\ie, identical to the TP of ground-truth image), it will bring positive impact on the estimated HR image, as we shown in Fig.~\ref{fig:comparison_TP}. Otherwise, if the categorical information in the extracted TP is incorrect, the SR result can be much damaged (please refer to Section~\ref{secdis} for such failure cases). Therefore, we use the TP extracted from the ground-truth HR image to supervise the learning of our TPGSR network. Denote by $t_L$ and $t_{H}$ the TP extracted from LR image $I_L$ and HR ground-truth image $I_H$, respectively, we use the $L_1$ norm distance $|t_H- t_L|$ and the KL divergence $D_{KL}(t_L ||t_H)$ to measure the similarity between $t_L$ and $t_{H}$. 
With the text priors $t_L, t_H \in R^{L \times |A|}$ of the pair of LR and HR images, the $D_{KL}(t_L ||t_H)$ can be calculated as follows:

\begin{equation}
    D_{KL}(t_{L}||t_{H}) = \sum_{i=1}^{L~}\sum_{j=1}^{|A|}t_H^{ij}\ln{\frac{t_H^{ij} + \epsilon}{t_L^{ij} + \epsilon}},
\label{equTP}
\end{equation}
where $t_L^{ij}$ and $t_H^{ij}$ denote the element in $i$th position and $j$th dimension in $t_{L}$ and $t_{H}$. $\epsilon$ is a small positive number to avoid numeric error in division and logarithm.
Together with $L_{S}$, the overall loss function for a single-stage TPGSR can is follows:
\begin{equation}
\begin{array}{l}
    L = L_{S} + \alpha~|t_{H} - t_{L}| + \beta~D_{KL}(t_{L}||t_{H}),
\end{array}
\label{equloss}
\end{equation}
where $\alpha$ and $\beta$ are the balancing parameters. 

For the multi-stage TPGSR learning, the loss for each stage, denote by $L_i$, can be similarly defined as in Eq.~\ref{equloss}. Suppose there are $N$ stages in total, the overall loss is defined as follows:
\begin{equation}
\begin{array}{l}
   L_{mt} = \sum_{i=1}^{N}\lambda_{i}~L_i,
\label{equ:loss}
\end{array}
\end{equation}
where $\lambda_{i}$ balances the loss of each stage and $\sum_{i=1}^{N}\lambda_{i}=1$.

\section{Experiments}

\subsection{Implementation Details}\label{sec:imple}

We implement our TPGSR method in Pytorch. Adam is selected as our optimizer with momentum $0.9$. The batch size is set to $48$ and the model is trained for $500$ epochs with one NVIDIA RTX 2080Ti GPU. The TP Generator is selected as CRNN~\cite{shi2016end} pre-trained on SynthText \cite{gupta2016synthetic} and MJSynth \cite{Jaderberg14c}. In Eq.~\ref{equloss}, the weights $\alpha$ and $\beta$ are 
both simply set to $1$%
, while the $\epsilon$ in Eq.~\ref{equTP} is set to $10^{-6}$. The alphabet set $A$ includes mainly alphanumeric ($0$ to $9$ and 'a' to 'z') case-insensitive characters. Together with a blank label, $|A|$ (\ie, the size of $A$), has $37$ categories in total. 
%
For dealing with the out-of-category cases, we assign all out-of-category characters with blank label, and the reconstruction of these characters mainly depends on the SR Module.

For multi-stage TPGSR training, we adopt a well-trained single-stage model to initialize all stages and cut the gradient across stages to speed up the training process to converge. The TP Generator are non-shared while the SR Module are shared cross stages. As in previous multi-stage learning methods~\cite{xie2019multispectral}, higher weight is assigned to the loss on the last stage, and the other stages are assigned with smaller weights on loss. In particular, we use a 3-stage TPGSR. The parameters $\lambda_i$ in Eq.~\ref{equ:loss} are set as $\lambda_1=\frac{1}{4}$, $\lambda_2=\frac{1}{4}$ and $\lambda_3=\frac{1}{2}$.

\begin{table*}[!htb]
\setlength{\abovecaptionskip}{0.2cm}
\captionsetup[subtable]{position=b}
\begin{subtable}[t][][b]{.5\linewidth}
\centering
\begin{tabular}{l||c|c}
\hline
Approach & Tuned & $\text{ACC}$\\\hline
TSRN~\cite{wang2020scene} & - &  41.4\%  \\
TPGSR-TSRN & $\times$ & 44.5\% \\
TPGSR-TSRN & $\checkmark$ &  \textbf{49.8\%}  \\\hline
HR & - &  72.3\%  \\\hline
\end{tabular}
\vspace{1\baselineskip}
\caption{Tuning the TP Generator.}
\label{table:ablationTP}
\end{subtable}%
\begin{subtable}[t][][b]{.5\linewidth}
\centering
\begin{tabular}{c|c|c|c|c}\hline
\textit{N} & E & M & H & ACC \\\hline
1 & 61.0\% & 49.9\% & 36.7\% & 49.8\%\\
2 & 62.2\% & 51.3\% & 37.4\% & 50.9\% \\
3 & 63.1\% & 52.0\% & 38.6\% & 51.8\%\\
4 & 63.7\% & 53.3\% & \textbf{39.4\%} & 52.6\%\\
5 & \textbf{64.3\%} & \textbf{54.2\%} & 39.2\% & \textbf{53.1\%}\\\hline
\end{tabular}
\caption{Ablation on different stage settings.}
\label{table:stage_settings}
\end{subtable} 
\begin{subtable}[t][][b]{.5\linewidth}
\vspace{1\baselineskip}
\centering
\begin{tabular}{c||c|c|c}
\hline
stage(\textit{N}) & TPG & SR & $\text{ACC}$ \\\hline
1 & \checkmark & $\times$ & 49.8\% \\\hline
2 & $\times$ & \checkmark & 50.9\% \\
3 & $\times$ & \checkmark & \textbf{51.8\%}  \\\hline
2 & $\times$ & $\times$ & 50.2\% \\
3 & $\times$ & $\times$ & 51.5\%  \\\hline
3 & \checkmark & \checkmark & 49.2\% \\\hline
3 & \checkmark & $\times$ & 49.2\%  \\\hline
\end{tabular}
\caption{Ablation on different sharing strategies.}
\label{table:AblationOnSharing}
\end{subtable} 
\begin{subtable}[t][][b]{.5\linewidth}
\vspace{3\baselineskip}
\centering
\begin{tabular}{c||cc|cc}
\hline
~ & \multicolumn{2}{|c}{$\text{ACC}$} & \multicolumn{2}{|c}{$\text{ACC}_T$} \\\hline
\textit{N} & $I_{LR}$ & $I_{SR}$ & $I_{LR}$ & $I_{SR}$ \\\hline
1 & 26.8\% & 49.8\% & \textbf{45.3\%} & 50.5\%\\
2 & 26.8\% & 50.9\% & 43.6\% & 51.6\% \\
3 & 26.8\% & \textbf{51.8\%} & 43.1\% & \textbf{52.9\%}\\\hline
\end{tabular}
\vspace{1\baselineskip}
\caption{Ablation on the impact of SR.}
\label{table:tunedTPG}
\end{subtable}

\caption{Ablation studies on TPGSR. ’Tuned’ means whether the TP Generator is fine-tuned or not. ACC means the average recognition accuracy, while $\text{ACC}_T$ means the average recognition accuracy with tuned TP Generator. \textit{N} means the stage number. 'E', 'M' and 'H' in (b) namely mean the accuracies of 'Easy', 'Medium' and 'Hard' split in TextZoom. ’\checkmark’ in (c) means that the component shares weights in all stages, while ’×’ means that the components are trained separately. 'TPG' and 'SR' in (c) mean the TP Generator and SR module in TPGSR. $I_{LR}$ and $I_{SR}$ in (d) refer to using the LR image and generated SR images as input to the recognizer.}
\vspace{-0.3cm}
\end{table*}



\subsection{Datasets and Experiment Settings}
\label{Datasets}

\noindent\textbf{Datasets.} The TextZoom~\cite{wang2020scene}, ICDAR2015~\cite{karatzas2015icdar} and SVT~\cite{wang2011end} datasets are used to validate the effectiveness of our proposed TPGSR method. TextZoom consists of $21,740$ LR-HR text image pairs collected by lens zooming of the camera in real-world scenarios. The training set has $17,367$ pairs, while the test set is divided into three subsets based on the camera focal length, namely \emph{easy} ($1,619$ samples), \emph{medium} ($1,411$ samples) and \emph{hard} ($1,343$ samples). The dataset also provides the text label for each pair.

ICDAR2015 is a well-known scene text recognition dataset, which contains $2,077$ cropped text images from street view photos for testing. SVT is also a scene text recognition dataset, which contains $647$ testing text images. Each image has a $50$-word lexicon with it.

\noindent\textbf{Experiment settings.} Since there are real-world LR-HR image pairs in the TextZoom dataset, we first use it to train and evaluate the proposed TPGSR model. We then apply the trained model to ICDAR2015/SVT to test its generalization performance to other datasets. Considering the fact that most of the images in ICDAR2015 and SVT have good resolution and quality, while the TextZoom training data focus on LR images, we perform the generalization test only to the low quality images in ICDAR2015/SVT whose height is less than $16$ or the recognition score is less than $0.9$.

\subsection{Ablation Studies}
\label{secablation}

To better understand the proposed TPGSR model, in this section we conduct a series of ablation experiments on the selection of parameters in loss function, the selection of number of stages and whether the TP Generator should be fine-tuned in training. We also perform experiments to validate the effectiveness of SR Module in our TPGSR framework. We adopt TSRN~\cite{wang2020scene} as the SR Module in the experiments, and name our model as TPGSR-TSRN. All ablation experiments are performed on TextZoom and the recognition accuracies are evaluated with CRNN~\cite{shi2016end}.


\noindent\textbf{Impact of tuning the TP Generator.} The loss terms for the TP branch aim to fine-tune the TP Generator. To prove the significance of TP Generator tuning, we conduct experiments by fixing and tuning the TP Generator in a one-stage TPGSR model. The text recognition accuracies are shown in Table~\ref{table:ablationTP}. By fixing the TP Generator, we can enhance the SR image recognition by $3.1\%$ compared to the TSRN baseline ~\cite{wang2020scene}. By tuning the TP Generator during the training process, the recognition accuracy can be further improved from $44.5\%$ to $49.8\%$, achieving a performance gain of $5.3\%$. This clearly demonstrates the benefits of tuning the TP Generator to the SR text recognition task.



\noindent\textbf{Impact of multiple stages in TPGSR}. In addition to refining the TP Generator, recurrently inputting the estimated HR image into the TPGSR can also enhance the quality of TP since the SR Module can improve the estimated HR text image in each recurrence. To find out how well the multi-stage refinement can reach, we set the stage number $N=1,2,…,5$ and report the text recognition accuracy in Table~\ref{table:stage_settings}. We can see that the recognition accuracy increases with the increase of $N$; however, the margin of improvement decreases with \textit{N}. When $N = 5$, the accuracy of 'Hard' split begins to fall. Considering the balance between the model size and the performance gain, we set \textit{N} to 3 in our following experiments.

\begin{table*}[t]
\scriptsize
\centering

\begin{tabular}{l|cccc|cccc|cccc}
\hline
{Approach} &   \multicolumn{4}{|c|}{ASTER~\cite{shi2018aster,asterpytorch}} & \multicolumn{4}{|c|}{MORAN~\cite{luo2019moran,moranpytorch}} & \multicolumn{4}{|c}{CRNN~\cite{shi2016end,crnnpytorch}}\\\hline
~ &  easy & medium & hard & \textbf{average} & easy & medium & hard & \textbf{average} & easy & medium & hard & \textbf{average}\\\hline
BICUBIC&  64.7\% & 42.4\% & 31.2\% & 47.2\% & 60.6\% & 37.9\% & 30.8\% & 44.1\% & 36.4\% & 21.1\% & 21.1\% & 26.8\%\\\hline
SRCNN~\cite{dong2014learning,wang2020scene} & 69.4\% & 43.4\% & 33.0\% & 49.5\% & 63.2\% & 39.0\% & 30.2\% & 45.3\% & 38.7\% & 21.6\% & 20.9\% & 27.7\%\\
TPGSR-SRCNN &  72.9\% & 50.7\% & 34.7\% & 53.8\% & 67.7\% & 49.7\% & 32.8\% & 50.9\% & 47.0\% & 30.6\% & 24.7\% & 34.7\%\\\hline

SRResNet~\cite{ledig2017photo,wang2020scene} &  69.6\% & 47.6\% & 34.3\% & 51.3\% & 60.7\% & 42.9\% & 32.6\% & 46.3\% & 39.7\% & 27.6\% & 22.7\% & 30.6\%\\
TPGSR-SRResNet &  76.0\% & 58.8\% & 40.1\% & 59.1\% & 72.3\% & 54.9\% & 38.4\% & 56.0\% & 54.6\% & 41.2\% & 32.3\% & 43.3\%\\\hline

RDN~\cite{zhang2018residual,wang2020scene} &  70.0\% & 47.0\% & 34.0\% & 51.5\% & 61.7\% & 42.0\% & 31.6\% & 46.1\% & 41.6\% & 24.4\% & 23.5\% & 30.5\%\\
TPGSR-RDN &  72.6\% & 54.2\% & 37.2\% & 55.5\% & 67.8\% & 51.7\% & 36.0\% & 52.6\% & 53.0\% & 38.0\% & 27.7\% & 40.2\%\\\hline
TSRN~\cite{wang2020scene} &  75.1\% & 56.3\% & 40.1\% & 58.3\% & 70.1\% & 53.3\% & 37.9\% & 54.8\% & 52.5\% & 38.2\% & 31.4\% & 41.4\%\\
TPGSR-TSRN &  \textbf{78.9\%} & \textbf{62.7\%} & \textbf{44.5\%} & \textbf{62.8\%} & \textbf{74.9\%} & \textbf{60.5\%} & \textbf{44.1\%} & \textbf{60.5\%} & \textbf{63.1\%} & \textbf{52.0\%} & \textbf{38.6\%} & \textbf{51.8\%}\\\hline\hline
HR & 94.2\% & 87.7\% & 76.2\% & 86.4\% & 91.2\% & 85.3\% & 74.2\% & 83.9\% & 76.4\% & 75.1\% & 64.6\% & 72.2\%\\\hline\hline
\end{tabular}
\caption{SR text image recognition performance of competing STISR models on TextZoom. The recognition accuracies are evaluated by the officially released models of ASTER~\cite{shi2018aster}, MORAN~\cite{luo2019moran} and CRNN~\cite{shi2016end}.}
\label{table:ablationTextZoom}
\end{table*}

\noindent\textbf{Parameter sharing strategy.} To determine the best sharing strategies, we conduct experiments to test on both the TP Module and the SR Module. As shown in Table~\ref{table:AblationOnSharing}, we find that under different settings of stage number, the setting of non-shared TP Module shows significant performance improvement. However, when we use non-shared SR Module, little performance improvement in SR image recognition is achieved. Thus we use the settings of shared SR Module and non-shared TP Module in our multi-stage model.

\noindent\textbf{The effectiveness of SR in TPGSR.} Since one of the goals of STISR is to improve the text recognition performance by HR image recovery, it is necessary to check if the estimated SR images truly help the final text recognition task. To this end, we evaluate the TPGSR models with both fixed and tuned TP Generator by using LR and SR images as inputs. For multi-stage version, we test all the TP Generators and pick the best LR and SR results from all TP Generators. Note that models with tuned TP Generator and LR image as input is similar to directly fine-tuning the text recognition model on the LR images. The results are shown in Table~\ref{table:tunedTPG}. It can be seen that by tuning TP Generator on the LR images, the text recognition accuracy can be increased. However, the recognition accuracy can be improved more by using the SR text image. For example, at stage one, the recognition accuracy of LR images by using tuned TP Generator is $45.3\%$, while the accuracy of SR images even without fine-tuning the TP Generator will be $49.8\%$. If the tuned TP Generator is used to generate the SR text image, the text recognition performance can be further improved compared to the fixed TP Generator. The experiments and comparisons demonstrate the effectiveness of our SR Module in improving the final SR text recognition.

\subsection{Comparison with State-of-the-Arts}
As described in Section~\ref{SecArcTPGSR} and illustrated in Fig.~\ref{fig:TPB}, the SR block of most existing representative SISR and STISR models can be adopted in the SR Module of our TPGSR framework, resulting in a new TPGSR model. To verify the superiority of our TPGSR framework, we select several popular SISR models, including SRCNN~\cite{dong2014learning}, SRResNet~\cite{ledig2017photo},  RDN~\cite{zhang2018residual}, and the latest state-of-the-art STISR model TSRN~\cite{wang2020scene}, and embed their SR blocks into our TPGSR framework. The corresponding STISR models are called TPGSR-SRCNN, TPGSR-SRResNet, TPGSR-RDN, and TPGSR-TSRN, respectively. The TextZoom~\cite{wang2020scene}, ICDAR2015~\cite{karatzas2015icdar} and SVT~\cite{wang2011end} datasets are used to compare these models as well as their prototypes. For fair comparison, all the models are trained trained in TextZoom dataset with the same settings.

\begin{figure*}[t]
  \centering
  \includegraphics[width=\linewidth]{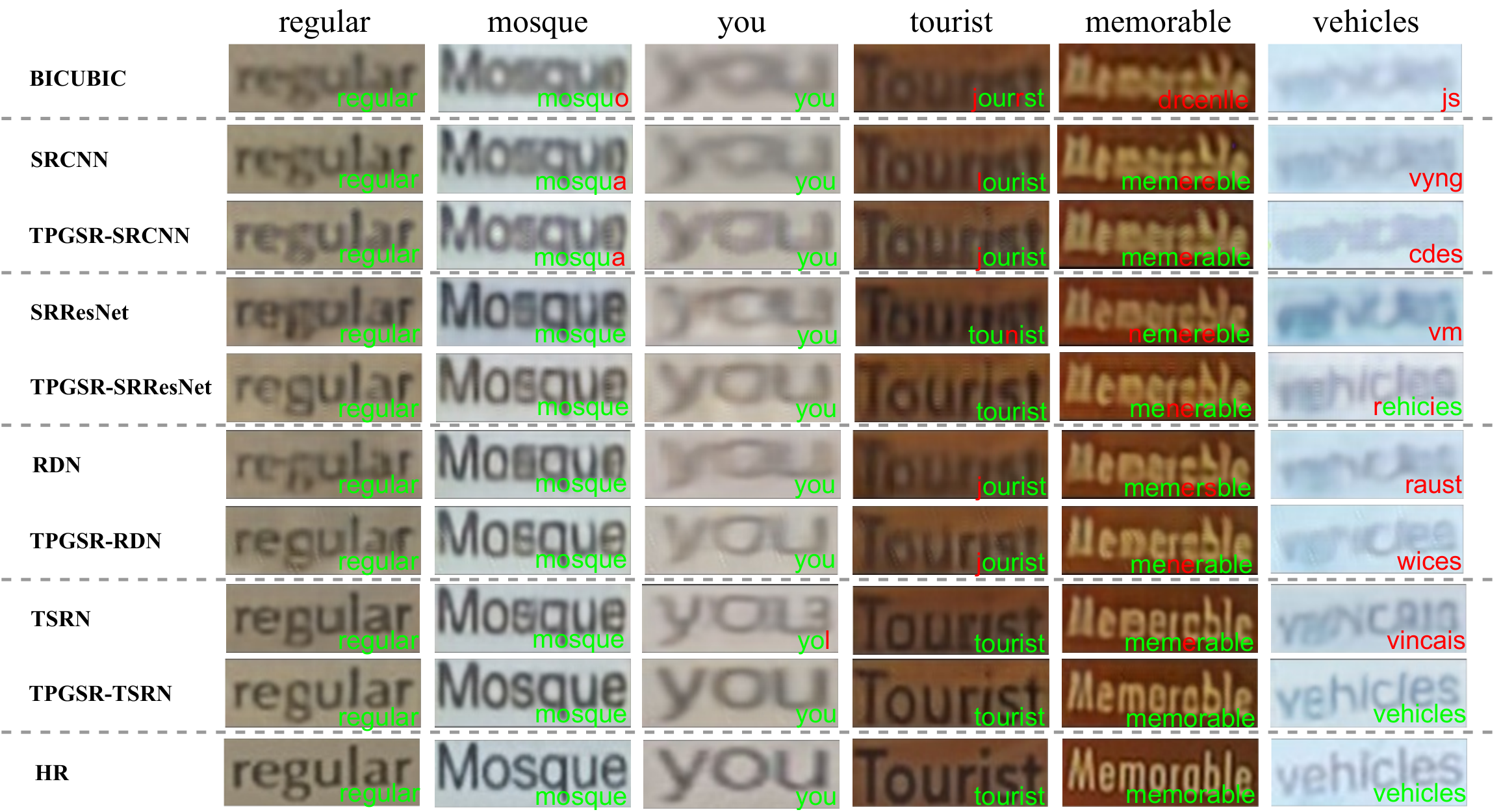}
  \caption{Visual comparison of competing STISR models on TextZoom. The word on the bottom-right corner of each image is the text recognition result, with correct characters or words in green and wrong in red.}
  \label{fig:vis_sr}
\vspace{-0.3cm}
\end{figure*}

\noindent\textbf{Results on TextZoom.}
The experimental results on TextZoom are shown in Table~\ref{table:ablationTextZoom}.
%
%
Here we present the text recognition accuracies on STISR results by using the official ASTER~\cite{shi2018aster}, MORAN~\cite{luo2019moran} and CRNN~\cite{shi2016end} text recognition models. In Fig.~\ref{fig:vis_sr}, we visualize the SR images by the competing models with the ground-truth text labels. 
From Table~\ref{table:ablationTextZoom} and Fig.~\ref{fig:vis_sr}, we can have the following findings. 


\begin{table}[t]
\begin{subtable}[t][][b]{\linewidth}
\centering
\footnotesize
\begin{tabular}{l||ccc|ccc}
\hline
\small{Dataset} & \multicolumn{3}{c}{ICDAR2015} & \multicolumn{3}{|c}{SVT}\\\hline
\small{No. of images} & \multicolumn{3}{c}{563}& \multicolumn{3}{|c}{104}\\\hline
\small{Approach} & SEED~\cite{qiao2020seed} & ASTER~\cite{shi2018aster} & CRNN~\cite{shi2016end} & SEED~\cite{qiao2020seed} & ASTER~\cite{shi2018aster} & CRNN~\cite{shi2016end}\\\hline
\small{Origin} & 54.0\% & 50.8\% & 21.5\% & 60.2\% & 50.8\% & 19.2\%\\\hline
\small{TSRN~\cite{wang2020scene}} & 52.6\% & 48.3\% &  24.5\% & 54.3\% & 48.3\% &  23.1\% \\
\small{TPGSR-TSRN} & \textbf{56.1\%} & \textbf{52.0\%} &  \textbf{27.1\%} & \textbf{61.1\%} & \textbf{52.0\%} &  \textbf{29.8\%}\\\hline
\end{tabular}
\end{subtable}
%
\caption{Text recognition accuracy on the low-quality images in ICDAR2015/SVT datasets by the TSRN and TPGSR-TSRN models trained on the TextZoom dataset.}
\label{table:ICDAR2015}

\end{table}

First, from Table~\ref{table:ablationTextZoom} we see that our TPGSR framework significantly improves the text recognition accuracy of all original SISR/STISR methods under all settings. This clearly validates the effectiveness of TP in guiding text image enhancement for recognition. 
%
Second, from Fig.~\ref{fig:vis_sr} we can see that with TPGSR, all SR models show clear improvement in text image recovery with more readable character stroke, resulting in correct text recognition. This also explains why our TPGSR can improve significantly the text recognition accuracy, as shown in Table~\ref{table:ablationTextZoom}.

\noindent\textbf{Generalization to other datasets.} As mentioned in Section \ref{Datasets}, to verify the generalization performance of our model trained on TextZoom to other datasets, we apply it to the low quality images (height $\leq16$ or recognition score $\leq0.9$) in ICDAR2015 and SVT. 
Overall, $563$ low quality images were selected from the $2,077$ testing images in ICDAR2015, and $104$ images were selected from the $647$ testing images in SVT. The STISR and text image recognition experiments are then performed on the $667$ low-quality images. Since TSRN~\cite{wang2020scene} is specially designed for text image SR and it performs much better than other SISR models, we only employ TSRN and TPGSR-TSRN in this experiment. The ASTER and CRNN text recognizers as well as stronger baseline SEED~\cite{qiao2020seed} are used.

The results are shown in Table~\ref{table:ICDAR2015}.  We can have the following findings. First, compared with the text recognition results using original images without SR, TSRN improves the performance when CRNN is used the text recognizer, but decreases the performance when ASTER or SEED is used as the recognizer. This implies that TSRN does not have stable cross-dataset generalization capability. Second, TPGSR-TSRN can consistently improve the performance over the original images in all three recognizer. This demonstrates that it has good generalization performance on cross-dataset test. Third, TPGSR-TSRN consistently outperforms TSRN under all settings.


\begin{table*}[t]
\small
\centering
\begin{tabular}{l|ccc|ccc|ccc}
\hline
Approach & \multicolumn{3}{c|}{Accuracy of ASTER~\cite{shi2018aster}} & \multicolumn{3}{c|}{PSNR} & \multicolumn{3}{c}{SSIM} \\\hline
~ & easy & medium & hard & easy & medium & hard & easy & medium & hard \\\hline
BICUBIC & 64.7\% & 42.4\% & 31.2\%& 22.35 & 18.98 & 19.39 & 0.7884 & 0.6254 & 0.6592 \\\hline

SRCNN~\cite{dong2014learning,wang2020scene} &69.4\% & 43.4\% & 33.0\%& 23.48 & 19.06 & 19.34 & 0.8379 & 0.6323 & 0.6791 \\
TPGSR-SRCNN & 72.9\% & 50.7\% & 34.7\% & 22.82 & 19.01 & 19.35 & 0.8232 & 0.6372 & 0.6798 \\\hline

SRResNet~\cite{ledig2017photo,wang2020scene} & 69.6\% & 47.6\% & 34.3\% & 23.48 & 19.06 & 19.34 & 0.8681 & 0.6406 & 0.6911 \\
TPGSR-SRResNet & 76.0\% & 58.8\% & 40.1\% & 22.47 & \textbf{19.09} & 19.59 & 0.8648 & 0.6359 & 0.7006 \\\hline

RDN~\cite{zhang2018residual,wang2020scene} & 70.0\% & 47.0\% & 34.0\% & 22.27 & 18.95 & 19.70 & 0.8249 & 0.6427 & 0.7113 \\
TPGSR-RDN & 72.6\% & 54.2\% & 37.2\% & 23.36 & 18.90 & 19.77 & 0.8512 & 0.6524 & 0.7155 \\\hline

$\text{TSRN}$~\cite{wang2020scene} & 75.1\% & 56.3\% & 40.1\% & \textbf{25.07} & 18.86 & 19.71 & \textbf{0.8897} & 0.6676 & 0.7302 \\
TPGSR-TSRN & \textbf{78.9\%} & \textbf{62.7\%} & \textbf{44.5\%} & 24.35 & 18.73 & \textbf{19.93} & 0.8860 & \textbf{0.6763} & \textbf{0.7487} \\

\hline
\end{tabular}
\vspace{0.3cm}
\caption{Recognition accuracy, PSNR (dB) and SSIM results of the competing STISR models on TextZoom. All experiment are conducted under the same settings.}
\label{table:ablationTextZoomSR}
\end{table*}

\begin{table}[t]
\small
\centering
\begin{tabular}{l|l|c|c}
\hline
Approach & TPG Backbone & Flops & ACC \\\hline\hline
TSRN~\cite{wang2020scene} w 5 SRBs & - & 0.91G & 41.4\% \\\hline
TSRN~\cite{wang2020scene} w 7 SRBs & - & 1.16G & 40.1\%\\\hline
TPGSR-TSRN $(N=1)$ & VGG~\cite{shi2016end} & 1.72G & 49.8\%\\\hline
\end{tabular}
\vspace{0.3cm}
\caption{Cost \vs performance. ACC means the average recognition accuracy. $N$ refers to the stage number of the TPGSR.}
\label{table:Cost}

\end{table}

\subsection{Discussions}
\label{secdis}

\noindent\textbf{Cost \vs performance.}
To further examine the value of our TPGSR, we compare the computational cost of our single-stage TPGSR with the TSRN~\cite{wang2020scene}. In Table~\ref{table:Cost}, the experiments and results show that straightly increasing the number of SRB blocks is not an effective way of gaining performance ($1.3\%$ accuracy drop). However, under our designed TPGSR network, the performance shows an improvement of $8.4\%$ compared to TSRN with 5-SRB. It is humble to conclude that generating text prior under our TPGSR framework is more valued than the additional cost it introduces.

\noindent\textbf{The PSNR/SSIM indices for STISR.}
In terms of STISR, better results of some objective metrics, \eg, PSNR and SSIM, do not always guarantee more accurate scene text estimation, and vice versa. Similar conclusion can be also found in \cite{wang2020scene} that PSNR and SSIM are not stable metrics for STISR. For better interpretation of this point, we adopt some popular SR models, including SRCNN~\cite{dong2014learning}, SRResNet~\cite{ledig2017photo}, RDN~\cite{zhang2018residual} and TSRN~\cite{wang2020scene}, and compare the results of these models before and after they are integrated into our TPGSR framework for joint optimization. The results are demonstrated in Table~\ref{table:ablationTextZoomSR}. We can see that the accuracy of all models boosts after they are embedded to our TPGSR framework.  In terms of the objective metrics, our best SR model under TPGSR consistently outperforms the competing methods for samples at the ``hard’’ difficulty level, since TP strengthens the power of the SR models in the challenging cases. In contrast, for samples at both ``easy’’ and ``medium’’ difficulty levels, joint optimization using our TPGSR does not always improve PSNR/SSIM results. This is because SR models trained without TP may suffer from over-fitting in easier cases. We could alleviate this issue by introducing TP loss between LR images and the predicted HR images into our objective function as a regularization term for joint optimization. In this way, as shown in Table~\ref{table:PSNR_REC}, we can obtain better perceptual quality and more accurate scene text image recognition results, which we believe are more valuable in real world applications than subtle rise in metrics such as PSNR and SSIM.

\begin{table}[t]
\centering
\begin{tabular}{c}

\begin{minipage}{0.8\textwidth}
      \includegraphics[width=\linewidth]{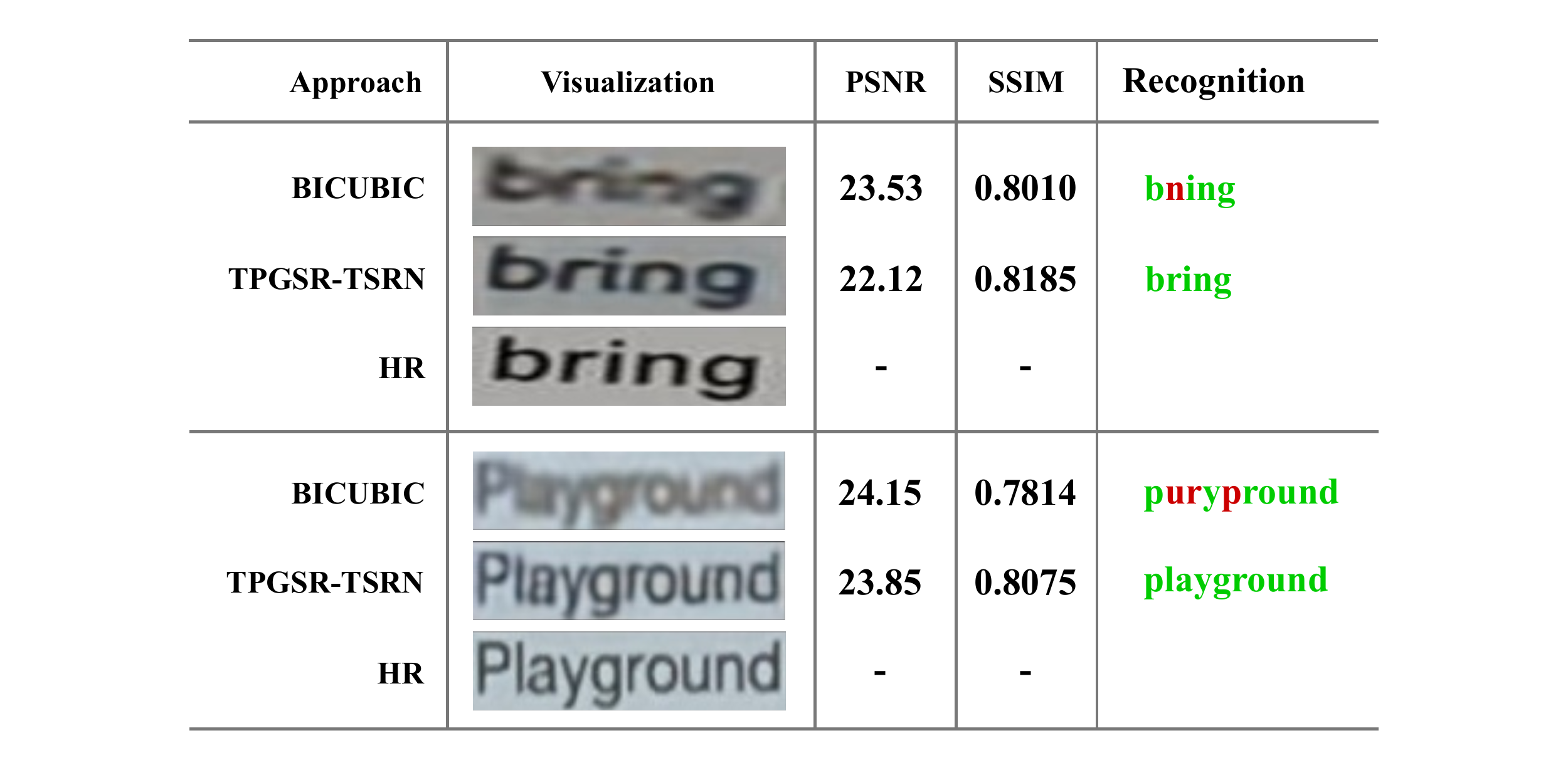}
    \end{minipage}
\end{tabular}
\vspace{0.3cm}
\caption{PSNR, SSIM and Recognition results of estimated HR images and real HR images.}
\label{table:PSNR_REC}
\end{table}

\begin{figure}[t]
\setlength{\abovecaptionskip}{0.2cm}
  \centering
  \includegraphics[width=0.7\linewidth]{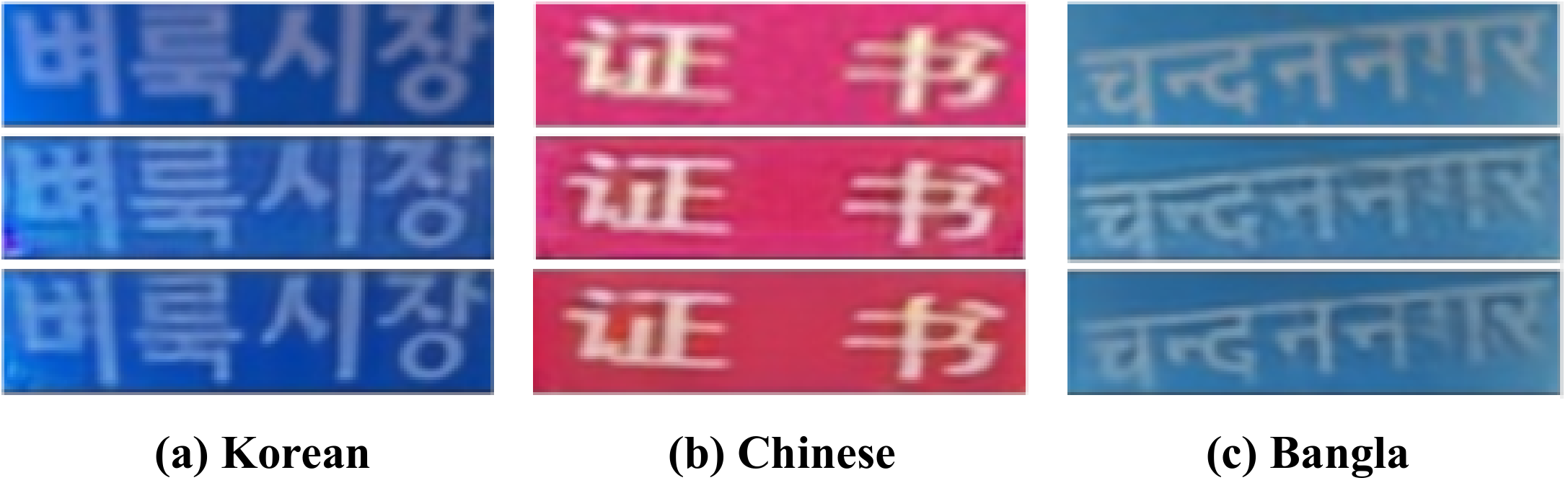}
  \caption{Examples of out-of-category text image SR in different languages. From top to bottom: the LR image and super-resolved HR images by TSRN~\cite{wang2020scene} and our TPGSR-TSRN. }
  \label{fig:out-of-category}
\vspace{-0.3cm}
\end{figure}

\noindent\textbf{Out-of-category analysis.} 
As mentioned in Section~\ref{sec:imple}, in our implementation, we assign the out-of-category characters with blank label. For such characters, the STISR results will mainly depend on the SR Module in our TPGSR network. To test the SR performance of our TPGSR model on images with out-of-category characters, we applied it to some text images in Korean, Chinese and Bangla picked from the ICDAR-MLT~\cite{nayef2017icdar2017} dataset. The results are shown in Fig.~\ref{fig:out-of-category}. We see that the reconstructed HR text images by our model show clearer appearance and contour than their LR counterparts. Compared with TSRN, TPGSR-TSRN demonstrates slightly 
better perceptual quality. 
The reason may be that categorical text prior serves as a regularization term in training SR Module to avoid over-fitting. Hence, the SR Module can still produce well-recovered scene text image with null guidance in inference stage.


\begin{figure}[t]
\setlength{\abovecaptionskip}{0.2cm}
  \centering
  \includegraphics[width=0.7\linewidth]{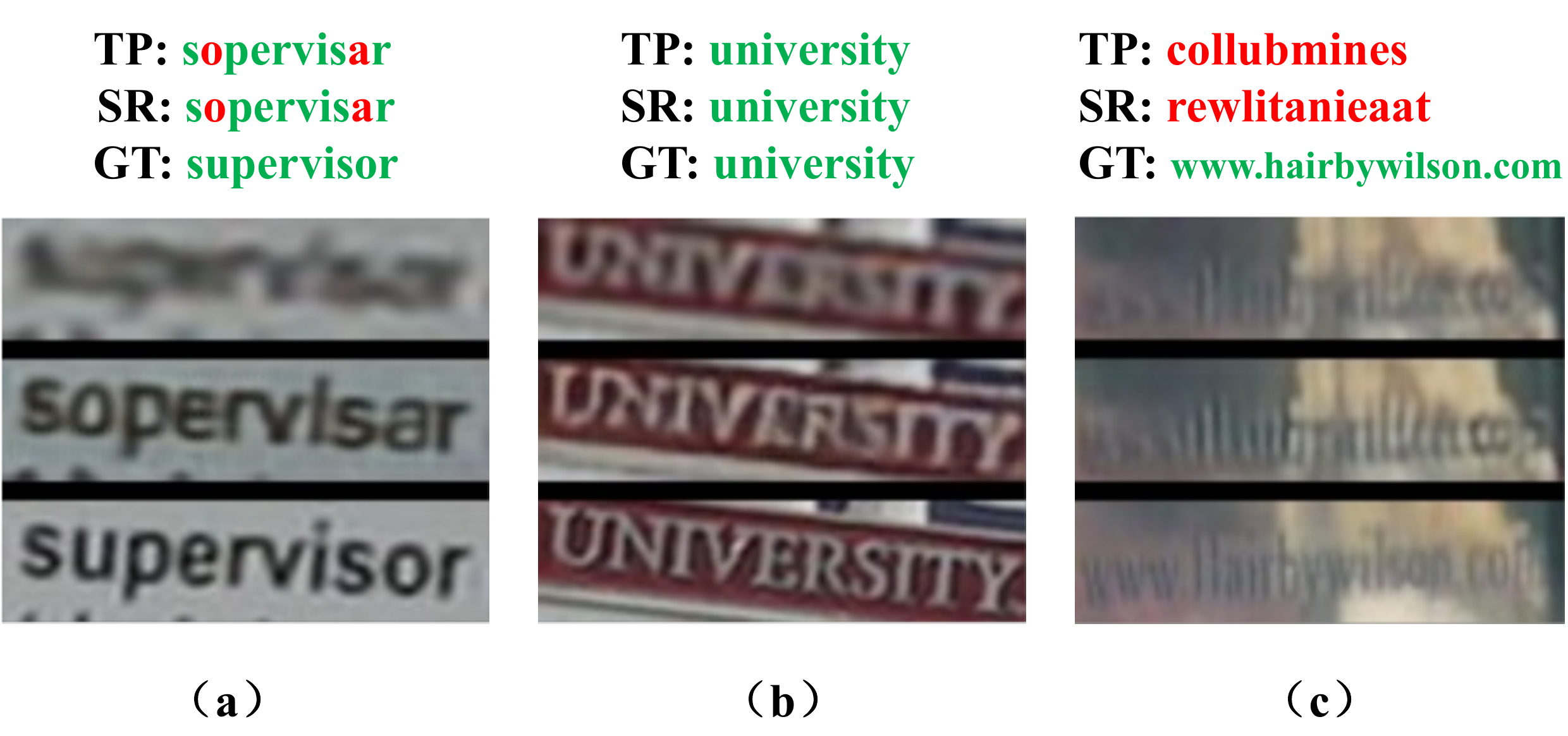}
  \caption{Examples of failure cases. (a) False TP guidance. (b) Multi-oriented text. (c) Extreme long text.}
  \label{fig:false_case}
\vspace{-0.3cm}
\end{figure}

\noindent\textbf{Failure case.} Though TPGSR can improve the visual quality of SR text images and boost the performance of text recognition, it still has some limitations, as shown in Fig.~\ref{fig:false_case}. First, the TP Generators tuned with LR samples are robust for most cases, while in some case it may produce false prior on the input LR image, and the output SR image may show incorrect character stroke, and causing false text recognition. Fig.~\ref{fig:false_case}(a) illustrates such a failure case. Second, if the text instance encounters certain rotation, the benefit brought by TPGSR will be weakened. Fig.~\ref{fig:false_case}(b) shows such an example. Though the recognition result is correct, the improvement on image quality is not significant. Third, for the cases with extremely long text instance, as shown in Fig.~\ref{fig:false_case}(c), the TP Generator may fail with significantly compressed outputs. In such case, the final output of TPGSR will suffer from text distortion with wrong text recognition.

To address the above issues, in the future we could consider to adopt more powerful TP Generators to provide more robust guidance, and design new TP guidance strategies for recovering multi-oriented scene text and curve text. In addition, the failures on long text can be alleviated by lengthening the width of input to the TP Generator.

\noindent\textbf{Recovering hieroglyphs~(e.g. Chinese).}
%
In this work, we focused on the real-world SR of English text images, for which there is a well-prepared benchmark dataset TextZoom. It is interesting to know whether our proposed method can be adopted for hieroglyphs such as Chinese. Here we perform some preliminary experiments to validate the feasibility. We train a multilingue recognition model using CRNN on the ICPR2018-MTWI Chinese and English dataset~\cite{icprmtwi} as our TP Generator. The overall alphabet contains $3,965$ characters, including the English and Chinese frequently-used set. Since there is no real-world benchmark dataset with LR-HR image pairs of Chinese characters, we synthesize LR-HR text image pairs by blurring and down-sampling the MTWI text images. We inherit the splits of MTWI as our training ($59,886$ samples) and testing ($4,838$ samples) set.  The model training and testing are conducted following the settings described in Section~\ref{sec:imple}.

\begin{figure}[t]
  \centering
  \includegraphics[width=0.8\linewidth]{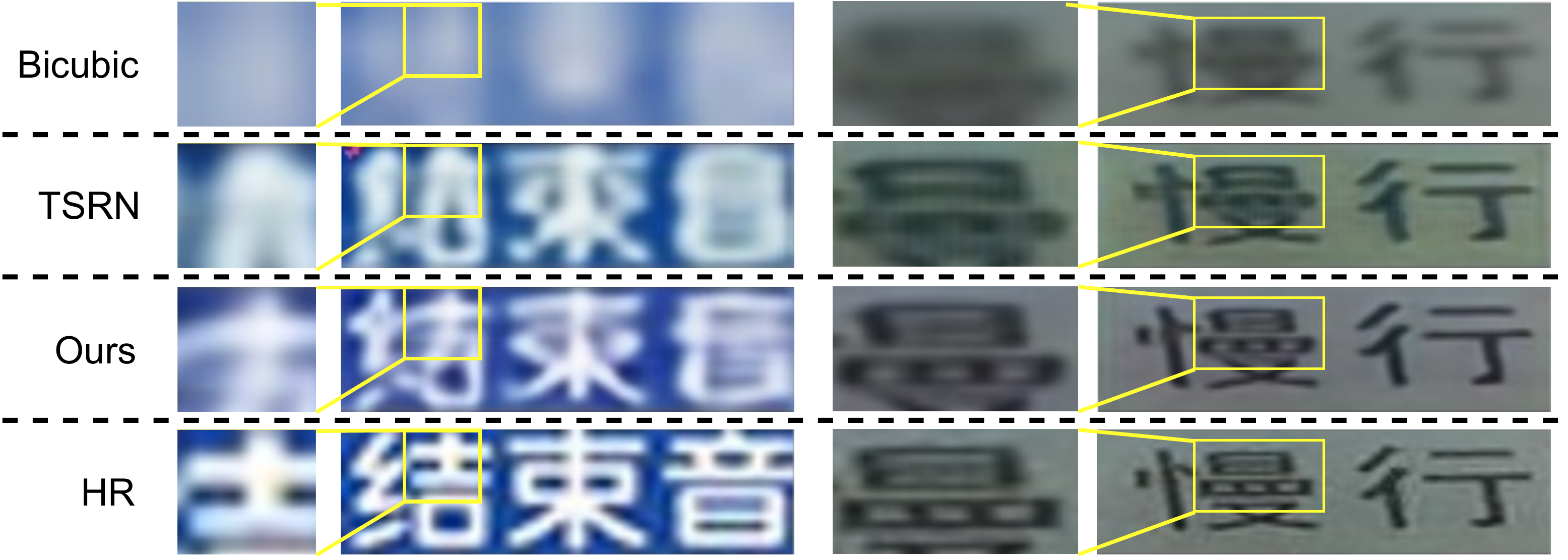}
  \caption{Application to the Chinese text recovery.}
  \label{fig:CHNSR}
\end{figure}

The SR text recognition results are $27.7\%$ (Bicubic), $41.1\%$ (TSRN~\cite{wang2020scene}), $42.7\%$ (TPGSR-TSRN) and $56.1\%$ (HR). From the results, we can observe that our TPGSR framework can still achieve $1.6\%$ accuracy gain over TSRN. Visualization on some Chinese characters can be seen in Fig.~\ref{fig:CHNSR}. One can see that our TPGSR can improve much the visual quality of SR results. Compared to TSRN~\cite{wang2020scene}, our TPGSR can better recover the text stroke on the samples. This preliminary experiment verifies that our TPGSR framework can be extended to hieroglyphs. More investigations and real-world dataset construction will be made in our future work.

\section{Conclusion}

In this paper, we presented a novel scene text image super-resolution framework, namely TPGSR, by introducing text prior (TP) to guide the text image super-resolution (SR) process. Considering the fact that text images have distinct text categorical information compared with those natural scene images, we integrated the TP features and image features to more effectively reconstruct the text characters. The enhanced text image can produce better TP in return, and therefore multi-stage TPGSR was employed to progressively improve the SR recovery of text images. Experiments on TextZoom benchmark and other datasets showed that TPGSR can clearly improve the visual quality and readability of low-resolution text images, especially for those hard cases, and consequently improve significantly the text recognition performance on them.

{\small
\bibliographystyle{ieee_fullname}

}



\end{document}